\documentclass{article}
\usepackage[utf8]{inputenc}
\usepackage{acl}

\usepackage{times}
\usepackage{latexsym}
\usepackage[T1]{fontenc}
\usepackage[utf8]{inputenc}
\usepackage{microtype}
\usepackage{hyperref}
\usepackage{amsmath,amsfonts}
\usepackage{algorithm}
\usepackage{algorithmic}
\usepackage{arydshln}
\usepackage{multirow,subfigure,booktabs}
\usepackage{caption}
\usepackage{amssymb}
\usepackage{graphicx}
\usepackage{needspace}

\title{
Entriever: Energy-based Retriever for Knowledge-Grounded Dialog Systems
}
\vspace{-1.0em}
\author{Yucheng Cai$^1$, Ke Li$^1$, Yi Huang$^{2,*}$, Junlan Feng$^2$, Zhijian Ou$^{1,}$
\thanks{$^*$Corresponding authors. This work is partly supported by the National Science and Technology Major Project (2023ZD0121401) and Guangxi Science and Technology Project (2022AC16002). The code for this paper is available at \url{https://github.com/thu-spmi/entriever}}\\
  $^1$
  SPMI Lab, EE Department, Tsinghua University, $^2$China Mobile Research Institute\\
  \texttt{{\{cyc22,ke-li24\}@mails.tsinghua.edu.cn}}, \texttt{ozj@tsinghua.edu.cn}\\
  \texttt{
  \{huangyi,fengjunlan\}@chinamobile.com}
  }
 \vspace{-1.0em} 

\newcommand{\modelname}{Entriever}

\newcommand{\comparemodel}{PL}

\begin{document}

\maketitle

\begin{abstract}

A retriever, which retrieves relevant knowledge pieces from a knowledge base given a context, is an important component in many natural language processing (NLP) tasks. 
Retrievers have been introduced in knowledge-grounded dialog systems to improve knowledge acquisition. 
In knowledge-grounded dialog systems, 
when conditioning on a given context, there may be multiple relevant and 
correlated knowledge pieces.
However, knowledge pieces are usually assumed to be conditionally independent in current retriever models. 
To address this issue, we propose \modelname{}, an energy-based retriever. \modelname{} directly models the 
candidate retrieval results as a whole instead of modeling the knowledge pieces separately, 
with the relevance score defined by an energy function. 
We explore various architectures of energy functions and different training methods for \modelname{}, and show that \modelname{} substantially outperforms the strong cross-encoder baseline in knowledge retrieval tasks.
Furthermore, we show that in semi-supervised training of knowledge-grounded dialog systems, \modelname{} enables  effective scoring of retrieved knowledge pieces 
and significantly improves end-to-end performance of dialog systems.

\end{abstract}

\vspace{-1.5em}

\section{Introduction}

\vspace{-0.5em}

Recently, with the development of large language models (LLMs), dialog systems have attracted increasing research interests. Although LLMs have shown an astonishing ability in open-domain question answering, they still often lack accuracy and make mistakes 
about certain facts in specific domains, such as customer services. 
Recent studies \cite{shuster2022blenderbot, izacard2022atlas, cai2023knowledge} 
have shown that the integration of knowledge retrieval into dialog systems can substantially enhance the precision of knowledge and mitigate the occurrence of hallucinations.
Therefore, knowledge retrieval is crucial to improve dialog systems, especially for those that require knowledge grounding.

Currently, two types of methods are prevalent for knowledge retrieval, statistical-based methods (like BM25) and dense retrieval methods (like DPR \cite{karpukhin2020dense}).
Both methods aim to find the most relevant piece of knowledge from a given knowledge base (KB).
However, when dealing with situations where multiple knowledge pieces from the KB might be relevant given certain context, which are common in real-life applications, both methods fail to account for the interrelationship among knowledge pieces and instead only model them separately. 
When a retrieval task requires the most relevant $n$ pieces of knowledge as a collective whole, these methods typically obtain the top $n$ results by relying on individual similarity scores. 
This ignores the relationship between these pieces, which could cause the retrieved pieces to contain repetitive information or miss important information. 

\begin{figure*}[t]
\centering
	\includegraphics[width=1\linewidth]{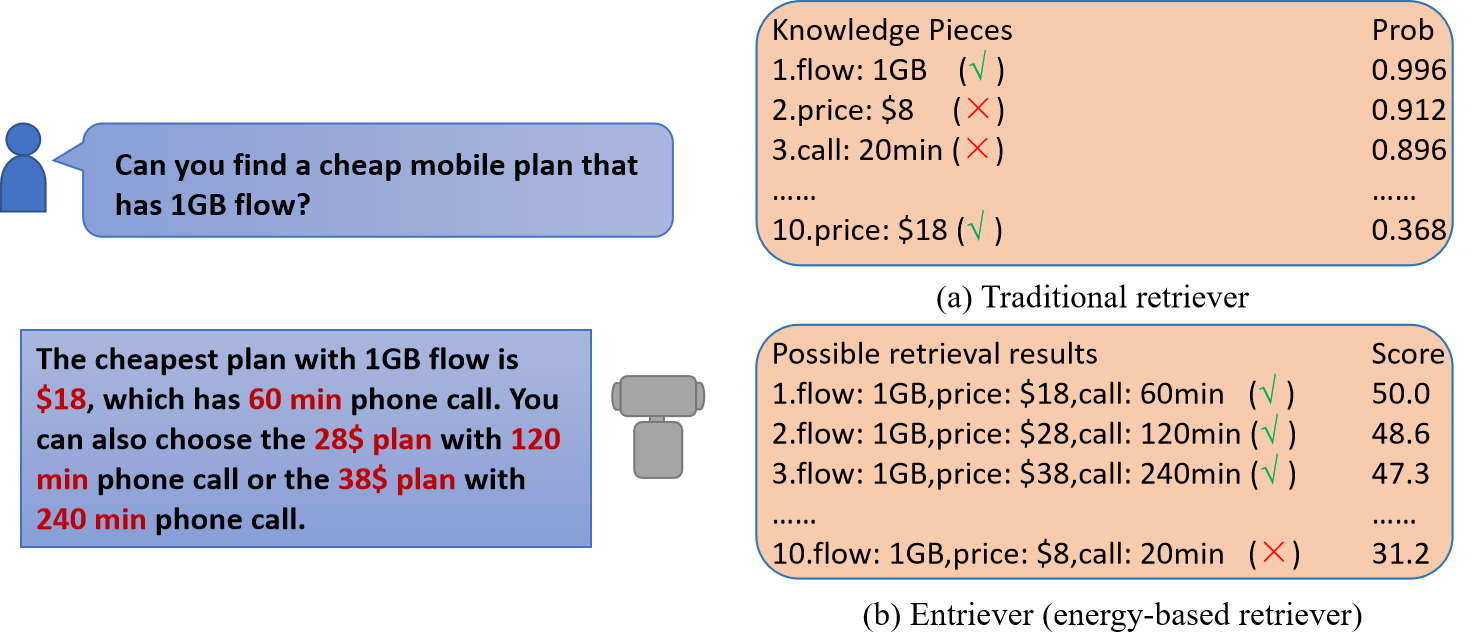}
 	\vspace{-1.0em}
	\caption{An illustration of the difference between (a) traditional retriever and (b) \modelname{} in the retrieval task for knowledge-grounded dialog systems. The traditional retriever judges the probability of each slots independently while \modelname{} assigns a score for each candidate retrieval result as a whole. 
    The traditional retriever ignores the interrelationship between knowledge pieces -- in this case the plan with 1GB flow should be at least \$18, and makes mistakes on the value of the price and call. 
    In contrast, \modelname{} can better model the interrelationships between knowledge pieces 
    and demonstrate better overall performance in the task.}
	\label{fig:entriever}
\end{figure*}

 To address the problems mentioned above, we propose using energy-based language models (ELMs) to model multiple knowledge pieces as a whole token sequence, rather than modeling the relevant knowledge pieces separately.
 A candidate retrieval result consists of multiple knowledge pieces.
 ELMs assign an energy score to each candidate result and use the score to distinguish positive candidates from negative ones, which is suitable for retrieval tasks. In previous research \cite{trf,learning_trf,bakhtin2021residual}, ELMs have been successfully used to calculate sentence scores in automatic speech recognition (ASR) and natural language generation (NLG). Different training strategies, such as noise contrastive estimate (NCE) and maximum likelihood estimate (MLE), have been explored \cite{learning_trf,DNCE,liu2023exploring}. However, to the best of our knowledge, 
our energy-based retriever, referred to as \modelname{}, is the first to use ELMs in retrieval tasks. We explore various MLE training approaches and find that the use of residual ELMs can greatly improve performance, shedding light on future work.

Moreover, 
note that the energy score of a candidate result is defined as the negative log probability up to an additive constant. The unnormalized nature of the energy score enables
the proposed \modelname{} to model the probability of a candidate result without the need to access the entire KB.
This feature is useful in building semi-supervised knowledge-grounded dialog systems. Semi-supervised knowledge-grounded dialog systems have 
seen significant progress recently \cite{deng2023regavae,cai2023knowledge}. 
These systems can make use of both labeled and unlabeled dialog data, reducing the reliance on costly manually labeled data and improving the efficiency of model training. 
However, it has been pointed out that the development of such systems can be difficult when the KB is not available in unlabeled data \cite{cai2023knowledge}.
This difficulty can be overcome by introducing \modelname{} that can model the probabilities of candidate results
without the need to access the entire KB. 
Experiments show that using the \modelname{} significantly improves the performance of semi-supervised knowledge-grounded dialog systems. 

Experiments are conducted on several knowledge-grounded dialog datasets, including MobileCS \cite{ou2022achallenge}, Camrest \cite{wen2017a}, In-Car \cite{eric2017key}, and Woz2.1 \cite{eric2019multiwoz}. 
We evaluate the performance of \modelname{} itself through the retrieval task and the performance gain that \modelname{} brings for semi-supervised dialog systems through the response generation task. 

In summary, the main contributions of this work are as follows.
\begin{itemize}
\item We propose to use an energy-based retriever (\modelname{}) to model 
each candidate retrieval result as a whole, which consists of multiple knowledge pieces, 
instead of modeling the knowledge pieces separately in knowledge-grounded dialog systems. 

\item We explore different architectures of energy functions and different training methods to train the \modelname{} and demonstrate 
the superiority of the proposed \modelname{} over previous methods. 

\item 
The proposed \modelname{} can model the retrieval probability without the need to access the entire KB,  which improves the performance of semi-supervised knowledge-grounded dialog system. 

 \end{itemize}

\vspace{-0.5em}

\section{Related Work}


\subsection{Knowledge Retrieval for Dialog Systems}


Recent research such as RAG \cite{lewis2020retrieval} and REALM \cite{guu2020realm} have introduced knowledge retrieval models into conditional generation, which greatly improves the quality of generated responses in knowledge-intensive tasks such as open-domain question answering and knowledge-grounded dialog systems. 
Retriever, which ranks relevant knowledge pieces from the KB given a context, is important for knowledge retrieval. Previous works use statistic-based retrievers (e.g. BM25 \cite{robertson2004bm25}) or neural network based retrievers 
(e.g. DPR \cite{karpukhin-etal-2020-dense}). There are several recent studies that improve over the original DPR retrievers. Re2G \cite{glass2022re2g} proposed to use a cross-encoder reranker to improve the performance over the dual-encoder retriever. Contriever \cite{izacard2022contriever} performed unsupervised pretraining using contrastive learning to improve the performance of the retriever.
RetroMAE \cite{xiao2022retromae} was pretrained with mask auto-encoding objective function to better capture the information of the whole sentence. Recently, LLM-based retrievers have gained much attention and achieve improvements over the traditional retrievers in some aspects \cite{shen2024retrieval,khramtsova2024leveraging}. However, all these works aim to retrieve the most relevant knowledge piece from the KB given certain searching context. Yet in real-life applications, multiple knowledge pieces from the KB might be relevant and helpful for response generation. Our work proposes to use an energy-based retriever (\modelname{}) to better model the inter-relationship among knowledge pieces instead of modeling them independently.
\vspace{-0.5em}
\subsection{Energy-based Language Models (ELMs)}
Energy-based language models (ELMs) parameterize an unnormalized 
distribution for natural sentences via an energy function, which can be very flexibly defined.
In previous studies, ELMs have shown promising performances in scoring for sentences in various applications such as computation of sentence likelihoods \cite{trf,learning_trf,wang2017language,DNCE,BinICASSP2018,gao2020integrating}, text generation \cite{deng2020residual}, language model pretraining \cite{electric}, calibrated natural language understanding \cite{joint_ebm} and calculating sentence scores in automatic speech recognition (ASR) \cite{liu2023exploring}.  Our work leverages the modeling flexibility of ELMs to model whether the ensemble of multiple knowledge pieces is suitable given a context.

There are two main training methods for ELMs, the maximum likelihood estimate (MLE) and the noise contrastive estimate (NCE) \cite{nce}. 
In this work, we mainly explore different architectures of energy functions and different sampling methods using MLE methods. In MLE, calculating gradients of the log likelihood usually resorts to Monte Carlo sampling methods. Two widely-used classes of sampling methods are importance sampling (IS) and Markov Chain Monte Carlo (MCMC) \cite{liu2001monte}. MCMC covers a range of specific algorithms and Metropolis independent sampling (MIS), where the proposed Markov move is generated independent of the previous state, is explored in this work. Meanwhile, residual ELM \cite{deng2020residual}, which models the ELM over a normalized model instead of modeling from scratch, is explored to study whether it can bring performance gain to the non-residual ELM. 

\subsection{Knowledge-Grounded Dialog Systems}
Knowledge-Grounded Dialog Systems aim 
to generate informative and meaningful responses
based on both conversation context and external
knowledge sources \cite{dinan2018wizard,kim2020sequential,zhao2020knowledge,li2022knowledge}. 
Semi-supervised knowledge-grounded dialog systems have seen significant progress recently \cite{li2020zero,paranjape2021hindsight,deng2023regavae,cai2023knowledge,cai20242nd}.
The use of semi-supervised training in knowledge-based dialog systems has been shown to greatly improve performance \cite{paranjape2021hindsight,deng2023regavae,cai2023knowledge}. In a semi-supervised knowledge-grounded dialog system, the knowledge required for response generation is not annotated in unlabeled data, and needs to be predicted by an inference model. 
An annoying difficulty in semi-supervised training is to accurately score the pseudo knowledge labels generated by the inference model on the unlabeled data. 
In previous efforts to build semi-supervised knowledge-grounded dialog systems \cite{cai2023knowledge}, researchers have to approximate the retrieval probability of the latent knowledge, using only the positive samples predicted by the inference model and ignoring the possible negative samples, due to that the KB is unavailable over unlabeled data. 
In contrast, \modelname{} directly models the retrieval probability of the latent knowledge as a whole. Our work explored using \modelname{} to better score the generated knowledge on the unlabeled data and significantly improved the performances of semi-supervised knowledge-grounded dialog system.

\vspace{-0.5em}

\section{Preliminary}

\vspace{-0.5em}

\label{sec:preliminary}

\subsection{Knowledge-Grounded Dialog Systems 
} 


\label{sec:model-defination}
Knowledge-grounded dialog systems 
retrieve relevant knowledge pieces given the dialog context and generate system response using the retrieved knowledge. Our settings of the knowledge-grounded dialog system is similar to  \cite{cai2023knowledge}.
Assume that we have a dialog 
with $T$ turns of user utterances and system responses, denoted by $u_{1},r_{1},\cdots,u_{T},r_{T}$ respectively. At turn $t$, based on the dialog context, the system queries a task-related KB to obtain relevant knowledge and generates appropriate responses.
The KB is made up of $N$ knowledge pieces\footnote{The form of knowledge pieces can be flexible, for examples, documents or items. In this paper, the knowledge pieces are mainly in the form of entities with attributes, or, say, slot-value pairs.}, denoted by 
$\{k^{1},k^{2},\cdots,k^{N}\}$ and the knowledge pieces that are relevant for the system to respond at turn $t$ are denoted by $\xi_{t}$. In knowledge-grounded dialog systems, the joint likelihood of the relevant knowledge pieces $\xi_{t}$ and the response $r_{t}$ given the context $c_{t}$ and user input $u_{t}$ is optimized at each turn $t$ in a dialog session. The likelihood is decomposed into a knowledge retrieval probability $p_\theta^{\text{ret}}$ and a response generation probability $p_\theta^{\text{gen}}$, as follows:
\begin{equation}
p_\theta(\xi_{t},r_{t}|c_{t},u_{t})
= p_\theta^{\text{ret}}(\xi_t|c_{t}, u_{t}) \times p_{\theta}^{\text{gen}}(r_t|c_{t}, u_{t}, \xi_{t})
\label{eq:decompose}
\end{equation}
The model parameters are collectively denoted by $\theta$, which can actually split into two parts $\theta=(\theta^{\text{ret}},\theta^{\text{gen}})$.

The retrieval model $p_\theta^{\text{ret}}$ is introduced 
to retrieve knowledge from the KB. 
Traditionally, knowledge pieces in the KB are modeled independently when multiple knowledge pieces are retrieved. 
Particularly, the knowledge piece $\xi_t$ necessary for turn $t$ is represented by $\xi_{t} \triangleq \xi_{t,1} \oplus \xi_{t,2} \oplus \cdots \oplus \xi_{t,N}$, where $\oplus$ denotes sequence concatenation. 
$\xi_{t,i} = k^{i}$ if the knowledge piece $k^{i}$ is relevant to the response $r_t$; otherwise, $\xi_{t,i}$ is set to be empty. 
Therefore, the retrieval probability can be written as follows:
\begin{align}
\vspace{-1.0em}
p_\theta^{\text{ret}}(\xi_t|c_{t}, u_{t}) = \prod \limits_{i=1}^{N} p_\theta^{\text{ret}}(\xi_{t,i}|c_{t}, u_{t})
\label{eq:retrieval}
\vspace{-0.5em}
\end{align}
which we refer to as traditional retriever. However, a drawback of traditional retrievers is that it ignores the interrelationship between the knowledge pieces and only models the knowledge pieces independently, since different knowledge pieces may contain similar or correlated information.

Traditional retrievers using Eq. (\ref{eq:retrieval}) can be implemented based on dual-encoders or cross-encoders.
In vertical domains where the KBs are small, cross-encoder retrievers are often used for their improved retrieval performance compared to dual-encoder retrievers.
In general domains where KBs are large and cross-encoders are computationally prohibitive, dual-encoder based retrievers are preferable for their reduced computation cost with fast k-nearest neighbor search library such as FAISS
\cite{karpukhin2020dense}.
In this work, the KBs in our experiments are small ones; thus a cross-encoder retriever based on BERT is used to realize $p_\theta^{\text{ret}}(\xi_{t,i}|c_{t}, u_{t})$, 
using $c_{t}\oplus u_{t} \oplus k^i$ as input,
as shown in Figure \ref{fig:retriever}(a).
 
\begin{figure}[t]
\centering
	\includegraphics[width=1\linewidth]{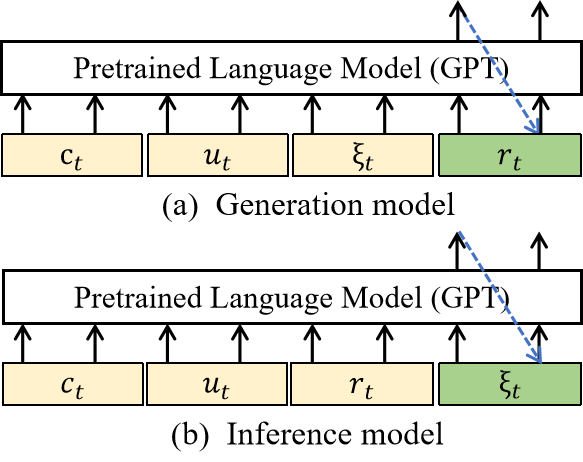}
 	\vspace{-1.0em}
	\caption{The models in semi-supervised training procedure: (a) the generation model, (b) the inference model. All of the variables ${c_t},{u_t},{\xi_t},{r_t}$ are represented by token sequences in our experiments.}
 	\vspace{-0.5em}
	\label{fig:models}
    \vspace{-1.0em}
\end{figure}

The generation probability ${p}_{\theta}^{\text{gen}}(r_{t} \mid c_{t}, u_{t}, \xi_{t})$ is instantiated with a GPT2 model using an autoregressive loss function, which is shown in Figure \ref{fig:models}(a): 
\begin{align}
&{p}_{\theta}^{\text{gen}}(r_{t} \mid c_{t}, u_{t}, \xi_{t}) \nonumber\\
&=\prod_{l=1}^{|r_{t}|}
p_{\theta}^{\text{gen}}(r_t^{(l)} \mid c_{t}, u_{t}, \xi_{t}, r_t^{(1)}, \ldots, r_t^{(l-1)}) \label{eq:gpt2_gen}
\end{align}
where $|\cdot|$ denotes the length in tokens, and $r_t^{(l)}$ the $l$-th token of the response $r_t$.

In training, the ground truth $\xi_t$, which is annotated in the dataset, is used to maximize the log probabilities in Eq. (\ref{eq:retrieval}) - (\ref{eq:gpt2_gen}). 
In testing, according to Eq. (\ref{eq:decompose}),  we firstly retrieve relevant slot-value pairs $\xi_t$; then, we generate $a_t$ and $r_t$, based on retrieved $\xi_t$. 
To be specific, to retrieve knowledge pieces using the cross-encoder  retriever in Eq. (\ref{eq:retrieval}), we threshold $p_\theta^{\text{ret}}(\xi_{t,i} = k^i |c_{t}, u_{t}), i=1,\cdots,N$, similar to \cite{cai2023knowledge}. 

\vspace{-0.5em}

\subsection{Semi-Supervision in Knowledge-Grounded Dialog Systems } \label{sec:semi-jsa}

\vspace{-0.5em}

Semi-Supervision aims to leverage both labeled and unlabeled data. Following \cite{cai2023knowledge}, our semi-supervised dialog systems use latent variable model and the joint stochastic approximation (JSA) algorithm \cite{ou2020joint} to optimize the latent variable model. 
As the knowledge pieces are annotated in labeled data and unavailable in unlabeled data, the relevant knowledge pieces 
$\xi_{1:T}$ are viewed as the latent variable for a dialog. Therefore, the generation model can be written as $p_\theta(\xi_{1:T}, r_{1:T}|u_{1:T})$ and the inference model can be written as 
$q_\phi(\xi_{1:T}|u_{1:T},r_{1:T})$ to approximate the true posterior $p_\theta(\xi_{1:T}|u_{1:T},r_{1:T})$. Both probabilities can be decomposed into the  turn , as pointed out in \cite{cai2022advancing} and shown in Figure \ref{fig:models}:

\vspace{-1.0em}

\begin{align}
\vspace{-1.0em}
p_\theta(\xi_{1:T}, r_{1:T}|u_{1:T}) =&\prod_{t=1}^T p_\theta(\xi_{t},r_{t}|c_{t},u_{t})
\label{eq:LS-1}\\
q_\phi(\xi_{1:T}|u_{1:T},r_{1:T})=&\prod_{t=1}^{T} q_\phi(\xi_{t}|c_{t},u_{t},r_{t}) 
\label{eq:posterior}
\vspace{-0.5em}
\end{align}

\vspace{-0.5em}

In supervised training, the ground truth knowledge $\xi_t$ is annotated in the dataset and thus can be directly used to maximize the probabilities in Eq. (\ref{eq:LS-1}) and Eq. (\ref{eq:posterior}). 
In semi-supervised training, the ground truth knowledge $\xi_t$ is not annotated for unlabeled data 
and should be inferred. Particularly, Metropolis independent sampling (MIS) is applied to draw samples from the true posterior $p_\theta(\xi_{t}|c_{t},u_{t},r_{t})$ using the inference model $q_\phi(\xi_{t}|c_{t},u_{t},r_{t})$ as a proposal. 
A recursive turn-level MIS sampler is used to sample $\xi_{1:T}$, as developed in \cite{cai2022advancing}. At each turn $t$, the MIS sampler works in a propose, accept or reject way, as follows:

1) Propose $\xi'_{t} \sim q_\phi(\xi_{t}|c_{t},u_{t},r_{t})$. 

2) Simulate $\eta \sim \text{Uniform}[0,1]$ and let
\begin{equation} \label{eq:tod-accept-reject}
\xi_{t} = \left\{
\begin{aligned}
&\xi'_{t}, &&\text{if}~\eta \le \min\left\lbrace 1, \frac{w(\xi'_{t})}{w(\tilde{\xi}_{t})} \right\rbrace \\ 
&\tilde{\xi}_t, &&\text{otherwise}
\end{aligned}
\right.
\end{equation}
where $\tilde{\xi}_t$ denotes the cached latent knowledge, and the importance weight $w(\xi_{t})$ between the target and the proposal distribution is defined as follows:
\begin{align}
w(\xi_{t}) 
&= \frac{p_\theta(\xi_t|c_{t}, u_t, r_t)}{q_\phi(\xi_{t}|c_{t},u_t,r_t)} \nonumber \\
&= \frac{p_\theta(\xi_t,r_t|c_{t}, u_t)}{q_\phi(\xi_{t}|c_{t},u_t,r_t)}\frac{1}{p_\theta(r_t|c_{t}, u_t)} \nonumber \\
& \propto \frac{p_\theta^{\text{ret}}(\xi_t|c_{t}, u_{t}) \times p_{\theta}^{\text{gen}}(r_t|c_{t}, u_{t}, \xi_{t}) }{q_\phi(\xi_{t}|c_{t},u_t,r_t)} \label{eq:tod-MIS-weight}
\end{align}
The term $p_\theta(r_t|c_{t}, u_t)$ is canceled out, since it appears in both the numerator and denominator of ${w(\xi'_{t})}/{w(\tilde{\xi}_{t})}$; and we only need to calculate the last line in Eq. (\ref{eq:tod-MIS-weight}) and use it as the importance weight.
The details of the JSA algorithm for training semi-supervised knowledge-grounded dialog systems are given in Appendix \ref{sec:jsa}.

Note that the retrieval probability $p_\theta^{\text{ret}}(\xi_t|c_{t}, u_{t})$ is needed to calculate the importance weight $w(\xi_{t})$ in Eq. (\ref{eq:tod-MIS-weight}). However, in previous works, using Eq. (\ref{eq:retrieval}) to calculate the retrieval probability requires to access the entire KB, which, however, is often not available for unlabeled data in semi-supervised knowledge grounded systems. To address this issue, we propose to use \modelname{} to calculate the retrieval probability of the latent knowledge $\xi_t$ as defined in Eq. (\ref{eq:ebm}).
\vspace{-0.5em}

\section{Method}

\vspace{-0.5em}

\begin{figure}[t]
\centering
	\includegraphics[width=1\linewidth]{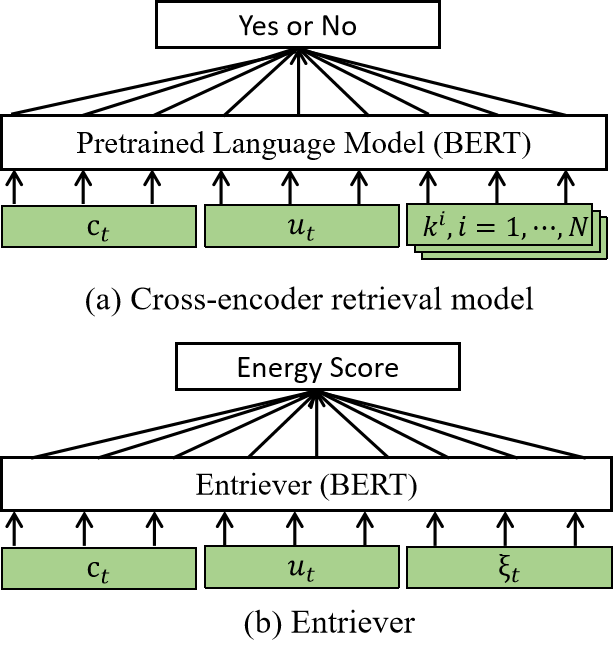}
 	\vspace{-1.5em}
	\caption{The architecture of retrieval models: (a) the cross-encoder retrieval model, which models the knowledge pieces independently, and (b) \modelname, which models the ensemble of relevant knowledge pieces. All of the variables ${c_t},{u_t},{\xi_t},{r_t},{k^i}$ are represented by token sequences.}
 	\vspace{-1.0em}
	\label{fig:retriever}
\end{figure}

As introduced in Section \ref{sec:preliminary}, there are two motivations to develop energy-based retriever (Entriever) - better modeling interdependencies between knowledge pieces and better enabling semi-supervised knowledge-grounded dialog systems. 
The former aims to model the candidate retrieval result as a whole, and the latter to model the retrieval probability without the need to access the entire knowledge base.
In Entriever, as shown in Figure \ref{fig:retriever}(b), the retrieval probability is defined by: 
\begin{align}
p_\theta^{\text{ret}}(\xi_t|c_{t}, u_{t}) &=\frac{{\exp(-U_\theta(c_{t}, u_{t}, \xi_{t}))}}{Z_\theta(c_{t}, u_{t})} \nonumber\\
&\propto {\exp(-U_\theta(c_{t}, u_{t}, \xi_{t}))} \label{eq:ebm}
\end{align}
where $U_\theta(c_{t}, u_{t}, \xi_{t})$ is the energy function. 
In this work, we initialize $U_\theta$ with BERT, similar to \cite{deng2020residual}.
$Z_\theta(c_{t},u_t)$ denotes the normalizing constant.


\subsection{Architecture of \modelname}


The architecture of the energy function 
$U_\theta(c_{t}, u_{t}, \xi_{t})$ in Eq. (\ref{eq:ebm}) can be very flexibly defined \cite{liu2023exploring}. In our work, a bi-directional text encoder (e.g., BERT) is used to encode the  input $c_{t}, u_{t}, \xi_{t}$ and we denote the encoder output (hidden vectors) by $\text{enc}_{\theta}(x)$. At position $i$, we have $\text{enc}_{\theta}(x)[i]$.
Then, the energy is defined as:
\begin{equation}
\label{eq:elm}
    U_\theta(c_{t}, u_{t}, \xi_{t}) = - \text{Linear} \left( \sum_{i=1}^{|x|} \text{enc}_\theta(x)[i] \right)
\end{equation}
where $\text{Linear}(\cdot)$ denotes a trainable linear layer whose output is a scalar and $x\triangleq c_{t}\oplus u_{t}\oplus \xi_{t}$ is the concatenation of the input sequence $c_{t}, u_{t}, \xi_{t}$. 

Orthogonal to the neural architecture used to define an energy function, we can define a residual form for an energy function, i.e., in the form of exponential tilting of a reference distribution \cite{learning_trf, deng2020residual}.
Specifically in our case, the retrieval probability can be defined as follows:
\begin{align}
p_\theta^{\text{ret}}(\xi_{t}|c_{t}, u_{t}) \propto {p^{\text{ref}}}(\xi_{t}|c_{t}, u_{t})\exp(-U_\theta(c_{t}, u_{t}, \xi_{t})) \label{eq:residual-ebm}
\end{align} 
where a reference distribution $p^{\text{ref}}(\xi_{t}|c_{t}, u_{t})$ is introduced. For simplicity, though with abuse of notation, we still use $U_\theta(c_{t}, u_{t}, \xi_{t})$ to denote the residual energy function, as for both non-residual and residual forms, we still use Eq. (\ref{eq:elm}) to realize $U_\theta$ and we will see in Section \ref{sec:training} that the formulas in model training share the same expressions.
The role of residual energy is to fit the difference between the target distribution and the reference distribution.

In this work, the reference distribution $p^{\text{ref}}(\xi_{t}|c_{t}, u_{t})$ is set to be the traditional retrieval distribution shown in Eq. (\ref{eq:retrieval}), which is usually closer to the target distribution than from uniform\footnote{The non-residual form in Eq. (\ref{eq:ebm}) can be viewed as a constrained subclass of the residual form in Eq. (\ref{eq:residual-ebm}), where the reference distribution is chosen to be uniform.}. 
Therefore, the residual Entriever only needs to learn the difference between the target distribution and the baseline distribution, which is easier to train.
Remarkably, as ${p^{\text{ref}}}(\xi_{t}|c_{t}, u_{t})$ is irrelevant to $\theta$, the residual Entriever can be optimized the same as the non-residual Entriever, which is introduced in Section \ref{sec:training}. In our experiments, we compare both forms of Entrievers and find that the residual Entriever reduces the training difficulty and brings substantial improvement to the overall performance of dialog systems.


\vspace{-0.5em}

\subsection{Training of \modelname{}}
\label{sec:training}
First, it should be noted that the formulas presented in Section \ref{sec:training} apply to both non-residual and residual forms of Entrievers, defined in Eq. (\ref{eq:ebm}) and Eq. (\ref{eq:residual-ebm}) respectively, unless otherwise specified.

MLE base model training of \modelname{} is to learn the energy function $ U_\theta(x)$, by using the negative log likelihood as the loss function:
\begin{equation}
\vspace{-0.5em}
\mathcal{J}_\theta=-\log p_\theta^{\text{ret}}(\xi_{t}|c_{t}, u_{t}). 
\end{equation}
The gradient of the loss function  
$\frac{\partial J_\theta(x)} {\partial \theta}$  can be derived as follows \cite{ou2024energy}:
\begin{align}
\label{eq:MLE}
\vspace{-0.5em}
    & \frac{\partial 
    \mathcal{J}_\theta(\xi_{t}|c_{t}, u_{t})} {\partial \theta} \nonumber \\ 
     =& - \frac{\partial U_\theta(c_{t}, u_{t}, \xi_{t})} {\partial \theta} + \mathbb{E}_{\xi_{t} \sim p_\theta^{\text{ret}}} \left[\frac{\partial U_\theta(c_{t}, u_{t}, \xi_{t})} {\partial \theta}\right]
\end{align}
The challenge in calculating the gradient in Eq. (\ref{eq:MLE}) is that calculating the second term as an expectation requires sampling from the unnormalized distribution $p_\theta^{\text{ret}}(\xi_{t}|c_{t}, u_{t})$, which is generally intractable. 
Similar to \cite{SNIS,liu2023exploring}, we compare two sampling approaches, Metropolis independence sampling (MIS) and important sampling (IS). Both approaches require a proposal distribution, which is set to be the traditional retrieval distribution in Eq. (\ref{eq:retrieval}) in this work and is denoted by $q(\xi_{t}|c_{t}, u_{t})$. Note that here we drop any parameters related to the proposal distribution, since it is always fixed during the training of \modelname{}. 

For the residual Entrievers in our experiments, we use the reference distribution $p^{\text{ref}}(\xi_{t}|c_{t}, u_{t})$ as the proposal distribution $q(\xi_{t}|c_{t}, u_{t})$ (i.e., both set to be the traditional retrieval distribution), we have the importance weight in the following simple form:
\begin{equation}
\frac{p_\theta^{\text{ret}}(\xi_{t}|c_{t}, u_{t})}{q(\xi_{t}|c_{t}, u_{t})} \propto \exp(U_\theta(c_{t}, u_{t}, \xi_{t})).
\end{equation}




\vspace{-0.5em}

\subsubsection{Importance Sampling (IS)}


Instead of directly sampling from the intractable distribution $p_\theta^{\text{ret}}(\xi_{t}|c_{t}, u_{t})$, importance sampling draw proposal samples from a tractable distribution $q(\xi_{t}|c_{t}, u_{t})$ \cite{liu2001monte}. 
The importance weight $\frac{p_\theta^{\text{ret}}(\xi_{t}|c_{t}, u_{t})}{q(\xi_{t}|c_{t}, u_{t})}$ is calculated and renormalization is taken to calculate the expectation. Specifically, to estimate the second term in Eq. (\ref{eq:MLE}), we can use the importance sampling method with the proposal distribution $q(\xi_{t}|c_{t}, u_{t})$: 

\vspace{-1.0em}
\begin{align}
&\mathbb{E}_{\xi_{t} \sim p_\theta^{\text{ret}}(\xi_{t} \mid c_{t}, u_{t})} \left[ \frac{\partial}{\partial \theta} {U}_{\theta}(c_{t}, u_{t}, \xi_{t}) \right] \nonumber\\
\approx & \frac{  \sum_{\xi_{t}}\frac{p_\theta^{\text{ret}}(\xi_{t} \mid c_{t}, u_{t})}{q(\xi_{t}|c_{t}, u_{t})} \frac{\partial{U}_{\theta}(c_{t}, u_{t}, \xi_{t})}{\partial \theta}}
{ \sum_{\xi_{t}}\frac{p_\theta^{\text{ret}}(\xi_{t} \mid c_{t}, u_{t})}{q(\xi_{t}|c_{t}, u_{t})}} ,\xi_{t} \sim q(\xi_{t}|c_{t}, u_{t})\nonumber
\label{eq:is}
\end{align}
where the samples  are from the proposal distribution $q(\xi_{t} \mid c_{t}, u_{t})$, which is set to be the traditional retrieval distribution in Eq. (\ref{eq:retrieval}). They can be trivially obtained by sampling from $N$ independent binary distributions.


\subsubsection{Metropolis Independence Sampling (MIS)}



Similar to the IS approach, the Metropolis Indepenence Sampling (MIS) approach draws proposal samples from the tractable proposal distribution $q(\xi_{t} \mid c_{t}, u_{t})$. Unlike the IS approach, which uses renormalization and weighted averaging techniques to estimate the expectation term, MIS uses Markov Chain Monte Carlo (MCMC) to obtain samples from the target distribution $p_\theta^{\text{ret}}(\xi_{t} \mid c_{t}, u_{t})$. MIS is a special case of Metropolis-Hasting \cite{liu2001monte} and has been applied for ELM in \cite{wang2017language,liu2023exploring}. 

In experiments, we run the Markov chain for $T$ steps. 
$\xi_t^{(0)}$ is randomly initialized.
At step $\tau=1,\cdots,T$, generate a proposal sample $\xi'_t$ from $q(\xi_{t} \mid c_{t}, u_{t})$, and accept $\xi_t^{(\tau)}=\xi'_t$ with probability 
\begin{equation}
\vspace{-1.0em}
\min \left\lbrace  1, \frac{p_\theta^{\text{ret}}(\xi'_{t} \mid c_{t}, u_{t})/q(\xi'_{t}|c_{t}, u_{t})}
{p_\theta^{\text{ret}}(\xi_t^{(\tau-1)} \mid c_{t}, u_{t})/q(\xi_t^{(\tau-1)}|c_{t}, u_{t})} \right\rbrace, \nonumber
\end{equation}
otherwise set $\xi_t^{(\tau)}=\xi_t^{(\tau-1)}$.
Then we can use the samples $\{\xi_t^{(1)},...,\xi_t^{(T)}\}$ to approximate the second term in Eq. (\ref{eq:MLE}) via Monte Carlo averaging:
\begin{equation}
\mathbb{E}_{\xi_{t} \sim p_\theta^{\text{ret}}(\xi_{t} \mid c_{t}, u_{t})}\left[\frac{\partial}{\partial \theta} {U}_{\theta}(c_{t}, u_{t}, \xi_{t})\right] 
\approx \frac{  \sum \limits_{\tau=1}^{T} \frac{\partial{U}_{\theta}(c_{t}, u_{t}, \xi_{t}^{(\tau)})}{\partial \theta}}
{T}  \nonumber \label{eq:mis}
\end{equation}


\subsection{Using \modelname{} to retrieve knowledge}

\subsubsection{Testing retrieval capability of \modelname{}}

During testing, for \modelname{} in Eq. (\ref{eq:ebm}), the retrieval candidate with the highest score is taken as the final retrieval result. 
The retrieval task in the  knowledge grounded dialog system aims at retrieving the complete set of useful knowledge pieces from the KB given the context. Therefore, a KB with $N$ knowledge pieces will yield $2^N$ possible combination of retrieval candidates, which is impossible to enumerate. 
Therefore, some traditional retriever (as defined in Eq. (\ref{eq:retrieval})) is firstly used as a proposal\footnote{In this work, we use the cross-encoder retriever for the proposal, but dual-encoders can be used as well.}. 
Only the $K$ retrieval candidates with the highest retrieval probabilities proposed from the traditional retriever are scored by \modelname{}.
The $2^N$ possible retrieval results are defined according to the traditional retriever,
which does not consider dependency among different knowledge pieces. 
Therefore, we can apply the Viterbi algorithm \cite{forney1973viterbi} to find the top-$K$ retrieval candidates from the traditional retriever,
instead of enumerating the probability of all $2^N$ possible combination of retrieval results to reduce computational complexity. We perform an ablation study to study the effect of $K$ on the retrieval results in Table \ref{tab:sample-num}. 

The pseudocode of the used Viterbi algorithm is shown in Algorithm \ref{alg:vertibi}.
It runs like a beam search, with beam width $K$. The index of knowledge pieces ($1,2,\cdots,N$) can be viewed as time steps, and at each step, there are two possible (step dependent) states to select from (select or not select). Each retrieval result is a knowledge piece subset, which can be viewed as a path through the $2 \times N$ lattice. The Viterbi algorithm can be applied to find the top-$K$ paths ($\eta_{1},\eta_{2}, \cdots,\eta_K$) traversing the lattice with their probabilities $p_{\eta_1},p_{\eta_2}, \cdots, p_{\eta_K}$.

\begin{algorithm}[t]
	\caption{The Viterbi algorithm to find the top-$K$ retrieval candidates from the traditional retriever}
 	\label{alg:vertibi}
	\begin{algorithmic}
	    \REQUIRE 
	    $N$ knowledge pieces $k^{1},k^{2}, \cdots, k^{N}$ with corresponding retrieval probabilities $p_{1},p_{2}, \cdots, p_{N}$ calculated by the traditional retriever Eq. (\ref{eq:retrieval}).
	    \STATE Initialize the retrieval knowledge piece subsets $\eta_{1},\eta_{2}, \cdots, \eta_{K}$ to empty;
	    \STATE Initialize the retrieval probabilities $p_{\eta_1},p_{\eta_2}, \cdots, p_{\eta_K}$ to be 1;
        \FOR{$i$ =1 to $N$}
        	\STATE Consider to expand the $K$ knowledge piece subsets by adding $k^i$ or not. If $k^i$ is selected and added to some $\eta_j, j=1,\cdots,K$, the probability $p_{\eta_j}$ is multiplied with $p_i$; otherwise, $p_{\eta_j}$ is multiplied with $1-p_i$;
        \STATE Calculate the possible $2 K$ probabilities for the subset expansion:
            $p_{\eta_1}*p_{i},p_{\eta_1}*(1-p_{i}),p_{\eta_2}*p_{i},p_{\eta_2}*(1-p_{i}), \cdots, p_{\eta_K}*p_{i},p_{\eta_K}*(1-p_{i})$;
            \STATE Select the top-$K$ results from the $2K$ results based on their probabilities, update $\eta_{1},\eta_{2}, \cdots, \eta_{K}$;
        \ENDFOR
        
        \RETURN The top-$K$ retrieval results $\eta_{1},\eta_{2}, \cdots, \eta_{K}$ with their retrieval probabilities.
	\end{algorithmic}
\end{algorithm}

\subsubsection{Leveraging \modelname{} in Semi-Supervised Knowledge-Grounded Dialog Systems}


In semi-supervised training of knowledge-grounded dialog systems, we need to calculate the importance weight $w(\xi_{t})$ in Eq. (\ref{eq:tod-MIS-weight}) in order to properly filter the generated pseudo labels for unlabeled data. 
This involves calculating the retrieval probability $p_\theta^{\text{ret}}(\xi_t|c_{t}, u_{t})$ for the pseudo labels $\xi_t$, generated from the inference model.
However, in unlabeled data such as customer service logs, KBs are often unavailable. 
This poses a significant challenge to the traditional retriever, which calculates the retrieval probability based on the entire KB by Eq. (\ref{eq:retrieval}). 
In contrast, the proposed \modelname{} can directly calculate the retrieval probability without the need to access the entire KB by Eq. (\ref{eq:ebm}).
Note that in semi-supervised experiments, since the KB is unavailable for unlabeled data, we only use the non-residual form of Entriever, i.e. Eq. (\ref{eq:ebm}).
In this setting, the unknown normalizing constant $Z_\theta(c_{t}, u_t)$ is canceled out, since it appears in both the numerator and denominator of ${w(\xi'_{t})}/{w(\tilde{\xi}_{t})}$ in Eq. (\ref{eq:tod-accept-reject}); and we can calculate the importance weight as follows, for $\xi_t$ generated from the inference model:
\begin{equation}
\vspace{-0.5em}
w(\xi_{t}) \propto \frac{\exp(-U_\theta(c_{t}, u_{t}, \xi_{t})) \times p_{\theta}^{\text{gen}}(r_t|c_{t}, u_{t}, \xi_{t}) }{q_\phi(\xi_{t}|c_{t},u_t,r_t)} \label{eq:entriever-importance-weight}
\end{equation}

The two-stage training of semi-supervised is detailed in Appendix \ref{sec:jsa}. The first stage is supervised pre-training of the retrieval model $p_\theta^{\text{ret}}$, the generation model $p_\theta^{\text{gen}}$, and the inference model $q_\phi$ on labeled data. 
In the second stage, the retriever is frozen, and only the generation model $p_\theta^{\text{gen}}$ and the inference model $q_\phi$ are further trained on the mix of labeled and unlabeled dialogs.

\begin{table*}[t]
\vspace{-1.0em}
    \caption{Results on knowledge retrieval task for the MobileCS, Camrest, In-Car, and Woz2.1 datasets. Joint-acc, Inform, and F1-score are reported. 
    Residual Entrievers are used and trained with different methods (MIS and IS).
    }
	\centering
	\resizebox{1\linewidth}{!}{
			\begin{tabular}{ c ccc ccc ccc ccc}
				\toprule
				\multirow{2}{*}{Method}  &\multicolumn{3}{c}{MobileCS}&\multicolumn{3}{c}{Camrest}
                &\multicolumn{3}{c}{In-Car}
                &\multicolumn{3}{c}{Woz2.1}\\
                  \cmidrule(lr){2-4} \cmidrule(lr){5-7}
                  \cmidrule(lr){8-10}
                  \cmidrule(lr){11-13}
                  &Joint-acc
    &Inform  &F1 &Joint-acc
    &Inform  &F1&Joint-acc
    &Inform  &F1&Joint-acc
    &Inform  &F1\\ 
				\midrule
  Cross-encoder 
  & 73.15  &35.95 & 0.589  & 81.38  &63.84 & 0.816 & 74.70  &42.16 & 0.870 & 75.00  &32.86 & 0.508  \\
   \hdashline

  \modelname{} (MIS)   &  76.67 & 39.81 &0.620 &  \textbf{83.17} & 68.05 &0.824&  \textbf{78.66} & 49.64 &\textbf{0.875}&  \textbf{80.24} & 43.78 &0.524 \\
  \modelname{} (IS)   &  \textbf{77.21}
   & \textbf{42.45} & \textbf{0.628}&  \textbf{83.17}
   & \textbf{68.28} & \textbf{0.825}&  78.51
   & \textbf{50.53} & \textbf{0.875}&  79.72
   & \textbf{45.02} & \textbf{0.530}\\

				\bottomrule
                \vspace{-1.0em}
		\end{tabular}}
		\label{tab:main-results}
\end{table*}

\vspace{-0.5em}

\section{Experiments}


\subsection{Experiment Settings and Baselines}

\begin{table}[t] 
    \caption{Comparison over the MobileCS dataset for different semi-supervision methods (pseudo labeling (PL) and JSA) 
    and whether \modelname{} is used or not during semi-supervised  training. Ratio means the ratio between the number of unlabeled dialogs and the number of labeled dialogs in training. The p-value denotes the significant test result for Combined score. The first colomun of p-value means whether 
    JSA + \modelname{} outperforms the PL methods, and the second colomun of p-value means whether the JSA + \modelname{} method significantly improves the JSA method.} 
	\centering
	\resizebox{1\linewidth}{!}{
			\begin{tabular}{cc cccc cccc}
				\toprule
				Ratio &Method  &Success &BLEU-4 &Combined &\multicolumn{2}{c}{p-value} \\ 
				\midrule
				\multirow{3}{*}{1:1} 
				&\comparemodel{}    & 87.5  &  8.853 & 105.21 &\multirow{3}{*}{0.025}&\multirow{3}{*}{0.013}\\
    & JSA  &  88.0 & 8.713 & 105.43 \\
				& JSA + \modelname{} 
                &   90.6 & 9.816     & \textbf{110.23} \\
				\midrule
				\multirow{3}{*}{2:1} 
				&\comparemodel{}  & 87.8  & 9.196 & 106.19 &\multirow{3}{*}{0.006}&\multirow{3}{*}{0.018}\\
    &JSA   & 88.7 & 9.490&   107.68 \\
				&JSA + \modelname{} 
                & 92.1 &  9.725   &   \textbf{111.55} \\
				\midrule
				\multirow{3}{*}{4:1} 
				&\comparemodel{}       & 88.5  & 9.341  & 107.18 
                &\multirow{3}{*}{0.049}&\multirow{3}{*}{0.088}\\
                &JSA   &90.9 & 9.398 &109.70 \\
				&JSA + \modelname{} 
                &92.8 & 9.554 &\textbf{111.91}\\
				\midrule
				\multirow{3}{*}{9:1} 
				&\comparemodel{}     & 89.4   & 9.532 & 108.46 &\multirow{3}{*}{ 0.083}&\multirow{3}{*}{ 0.192}\\
    &JSA  & 91.8 &9.677& 111.15\\
				&JSA + \modelname{} 
                & 93.0 & 9.627& \textbf{112.25}
\\
				\bottomrule
		\end{tabular}}
		\label{tab:compare-results}
\end{table}


Experiments are conducted on several dialog datasets: \textbf{(1) MobileCS dataset}, a real-life human-human dialog dataset, focuses on mobile customer service, released from the EMNLP 2022 SereTOD Challenge \cite{ou2022achallenge}. MobileCS contains a total of around 100K dialogs.
The labeled part was officially split into training/validation/test sets with 8,953/1014/955 dialogs, respectively. The remaining 87,933 dialogs are unlabeled.
\textbf{(2) CamRest dataset} \cite{wen2017a} focuses on dialogs in the restaurant domain, consisting of 676 dialogues. Each dialogue contains a KB. The average size of the KB is 22.5 triples. Following previous work, the dataset is split into training/validation/test sets with
406/135/135 dialogs.
\textbf{(3) In-Car Assistant dataset} \cite{eric2017key} comprises 3,031 dialogs spanning three domains: weather, navigation, and schedule. The average size of the KB for each dialogue is 62.3 triples. Following previous work, the dataset is split into training/validation/test sets with 2425/302/304 dialogs. 
\textbf{(4) Woz2.1 dataset} \cite{eric2019multiwoz} contains three domains: hotel, attraction and restaurant. The average size of the KB for each dialogue is 54.4 triples. Following \citep{ding2024retrieval}, the dataset is split into training/validation/test
sets with 1,839/117/141 dialogs.


For evaluation, we follow the scripts in \cite{cai2023knowledge} and \cite{ding2024retrieval}. 
We evaluate the knowledge retrieval ability of \modelname{} on all four dialog datasets. 
Three metrics, \emph{Joint Accuracy} (whether the whole knowledge in a dialog turn is accurate or not), \emph{Inform}  (whether the retriever provides all the key information for completing a dialog session), and \emph{F1} (the accuracy of the knowledge pieces retrieved), are reported.
To evaluate the improvement that \modelname{} brings to the semi-supervised knowledge-grounded dialog systems, experiments are taken on the MobileCS dataset, as only the MobileCS dataset contains unlabeled data. 
Two metrics, \emph{Success rate} and \emph{BLEU}, are used to evaluate the quality of the generated responses. 
\emph{Success rate} measures how often the system is able to provide all the entities and values requested by the user, which is crucial in performing a successful dialog. \emph{BLEU} is used to measure the fluency of the generated responses by analyzing the amount of n-gram overlap between the real responses and the generations. 
The overall performance of the semi-supervised knowledge-grounded dialog system is measured by \emph{Combined score}, which is \emph{Success} + 2*\emph{BLEU}, as in the original SereTOD challenge evaluation scripts \cite{liu2022information}. 

For the knowledge retrieval task, we select the most prevalent retriever, the dual-encoder retriever \cite{karpukhin-etal-2020-dense}, and the most competitive retriever, the cross-encoder retriever \cite{glass2022re2g,cai2023knowledge} (mostly used in the reranking tasks), as our baselines. For the semi-supervised knowledge-grounded dialog systems on MobileCS, several baselines are reported in the experiments. We implement \modelname{} upon the current state-of-the-art (SOTA) method JSA-KRTOD \cite{cai2023knowledge}. 

In our experiments, BERT \cite{devlin2019bert} is used to initialize the retrievers (including the dual-encoder baseline, cross-encoder baseline, and the Entriever) and GPT-2 \cite{radford2019gpt2} is used to initialize the response generator in the semi-supervised knowledge-grounded dialog systems following previous settings \cite{cai2023knowledge,ding2024retrieval}.
Hyper-parameters are chosen based on the development set, and evaluated on the test set.
\vspace{-0.5em}
\subsection{Main Results}
\begin{table}[t]
    \caption{Semi-supervised response generation results on the MobileCS dataset. Success, BLEU-4, and Combined score are reported. 
    }
    \vspace{-1.0em}
	\centering
	\resizebox{1\linewidth}{!}{
			\begin{tabular}{c c c c c}
				\toprule
    Method  &Success &BLEU-4 &Combined \\ 
				\midrule
  Baseline \cite{liu2022information}  & 31.5  &4.170  &39.84\\
   Passion \cite{lu2022passion} &43.2   &6.790  &56.78\\
   TJU-LMC \cite{yang2022discovering}& 68.9 &7.54  &83.98\\
PRIS \cite{zeng2022semi} & 78.9 &14.51  &107.92\\
JSA-KRTOD \cite{cai2023knowledge} & 91.8 &9.677& 111.15\\
JSA-KRTOD+\modelname{} (ours)& 93.0 & 9.627& \textbf{112.25}\\
				\bottomrule
		\end{tabular}}
		\label{tab:end2end}
\vspace{-0.5em}
\end{table}

\begin{table}[t]
    \caption{Knowledge retrieval capability on MobileCS for different model architectures and training methods. Joint-acc, Inform, and F1-score are reported. 
    }
    \vspace{-0.5em}
	\centering
	\resizebox{1\linewidth}{!}{
			\begin{tabular}{ l c c c c c}
				\toprule
				Setting &Joint-acc
    &Inform  &F1 \\ 
				\midrule
   Dual-encoder \cite{karpukhin-etal-2020-dense} &
   65.60 & 32.17  & 0.563 \\
  Cross-encoder \cite{cai2023knowledge}  & 73.15  &35.95 & 0.589\\
   \hdashline

   \modelname{} (Non-residual, MIS)    & 76.94&31.89 &  0.593\\
   \modelname{} (Non-residual, IS)    & 72.19 &32.22 & 0.596\\
  \modelname{} (Residual, MIS)   &  76.67 & 39.81 &0.620\\
  \modelname{} (Residual, IS)   &  \textbf{77.21}
   & \textbf{42.45} & \textbf{0.628}\\

				\bottomrule
		\end{tabular}}
		\label{tab:cross-dual}
        \vspace{-0.5em}
\end{table}

The experiments mainly explore the following research questions: \textbf{RQ1}: Whether \modelname{} can improve the knowledge retrieval performance? 
\textbf{RQ2}: Whether introducing \modelname{} can improve the overall performance of the semi-supervised knowledge-grounded dialog system?

As shown in Table \ref{tab:main-results}, \modelname{} greatly improves over the cross-encoder (current SOTA method) on all of the Joint-acc, Inform, and F1 metrics across the four datasets. Regardless of the training methods (MIS and IS), \modelname{} consistently outperforms the strong cross-encoder baseline. Based on these results, we can further discuss the reason for the improvement. The cross-encoder model (Figure \ref{fig:retriever}(a)) models the knowledge pieces in the KB independently given the context. In contrast, \modelname{} (Figure \ref{fig:retriever}(b)) models the collection of all relevant knowledge pieces 
given the dialog context.
In knowledge-grounded dialog systems, the interconnectivity and interdependence among relevant knowledge fragments are of great importance. 
Therefore, through the explicit modeling of such interrelationships, 
the retrieval results produced by \modelname{} tend to be more accurate as a whole, 
therefore achieving significantly higher scores on the Joint-acc and Inform metrics. 
These findings answer \textbf{RQ1} and show that \modelname{} can substantially improve the knowledge retrieval performance.


Considering the difference the importance sampling (IS) method and Metropolis independence sampling (MIS) method, there is no significant difference between the results. 
The sample size in IS and the markov steps in MIS are both set to be 12.
MIS and IS sampling methods perform equally well.
This is similar to the results of using ELMs in rescoring for speech recognition \cite{liu2023exploring}.

To answer \textbf{RQ2}, we conduct experiments on the semi-supervised knowledge-grounded dialog systems with \modelname{}. To systematically study the effect of the \modelname{}, different semi-supervised methods and label ratio are explored. 
As shown in Table \ref{tab:compare-results}, the introduction of \modelname{} substantially improves the overall performance (Combined Score) of the system regardless of the label ratio. Remarkably, the introduction of \modelname{} can greatly improve the Success rate metric for the dialog systems, indicating that the system's ability to provide the important knowledge is improved.
Moreover, the significant test results in Table \ref{tab:compare-results} show that almost in all settings, introducing \modelname{} can significantly improve the performances (p-value<0.1) over the original JSA method (filter the generated knowledge label with a less accurate knowledge retriever) and the pseudo labeling (PL) method (do not filter the generated knowledge label at all). 
Furthermore, as shown in Table \ref{tab:end2end}, our \modelname{} improves over the current SOTA semi-supervised method JSA-KRTOD \cite{cai2023knowledge} in MobileCS. These findings answer \textbf{RQ2} and show that introducing \modelname{} can improve the overall performance of semi-supervised knowledge-grounded dialog systems.

\vspace{-0.5em}

\subsection{Analysis and Ablation}
\label{sec:ablation}
\begin{table}[t]
    \caption{Ablation study on how the number of proposed candidate retrieval results ($K$) for \modelname{} to score influences the final test results over MobileCS.}
    \vspace{-0.5em}
	\centering
	\resizebox{1\linewidth}{!}{
			\begin{tabular}{ c c c c c c}
				\toprule
				Config  &Joint-acc
    &Inform &Precision &Recall &F1 \\ 
				\midrule
  $K=4$  & 76.02  &39.33 &0.7162& 0.5376 & 0.6142\\
   $K=8$    & 76.73 &40.70 & 0.7054 & 0.5580& 0.6231\\
   $K=16$   & \textbf{77.21}
   & 42.45 & 0.6855 & 0.5789 & \textbf{0.6277}\\
   $K=32$   &  76.79 & \textbf{42.60} & 0.6455& 0.6076 &0.6260\\

				\bottomrule
		\end{tabular}}
		\label{tab:sample-num}
\end{table}

\setlength{\aboverulesep}{0pt}
\setlength{\belowrulesep}{0pt}

\begin{table}[t] 
    \caption{The computational resource overhead when training and testing with different models on the MobileCS dataset. The table reports the time cost for one epoch of training on the training set and the complete inference on the test set, as well as the maximum GPU memory usage. The unit of time in the table is seconds, and the unit of GPU memory usage is megabytes (MB). For all the metrics, the smaller the numerical value, the better. 
    We use 3090 GPUs, and the training batch size is 8 and the batch size for inference is 32.}
      \vspace{-0.5em}
	\centering
	\resizebox{\linewidth}{!}{
	\begin{tabular}{ c | c c| c c }
				\toprule
				\multirow{2}{*}{Model}  &\multicolumn{2}{c|}{Training} 
     &  \multicolumn{2}{c}{Inference}  
    \\ &Time & GPU Memory &Time & GPU Memory\\
				\midrule
  Dual-encoder 
  &{905} & {16356} &  {34.89} & {3034}\\

   Cross-encoder  
   & 1469 & 21740 &65.58 & 3242  \\
   \modelname{} & 3803 & 24538
    & 170.91 & 3606\\
   
				\bottomrule
		\end{tabular}}
		\label{tab:computation}
    \vspace{-0.5em}
\end{table}

To further study the influence of using different architectures and training methods of the retrievers, an ablation study is conducted on the MobileCS dataset to evaluate the knowledge retrieval performance. As shown in Table \ref{tab:cross-dual}, the cross-encoder architecture greatly outperforms the commonly-used dual-encoder architecture, making it a strong baseline. For \modelname{}, the results show that the residual form of Entriever greatly improves the stability and performance of the training. Presumably, this is because that the residual form is built upon a trained cross-encoder retriever, which reduces the training burden and improves the training efficiency. 

We also conduct an ablation study to explore how the number of proposed candidate retrieval results ($K$) for \modelname{} to score
will affect the knowledge retrieval results, and the experiment results are shown in Table \ref{tab:sample-num}. 
From Table \ref{tab:sample-num}, it can be seen that the test results generally increase with the increase of $K$, when $K$ is relatively small, presumably because the oracle retrieval result is more likely to be covered. However, although continuously increasing $K$ can increase the possibility of providing the correct knowledge, more noisy samples are introduced as well. Moreover, the computational budgets increase linearly with $K$. As shown in Table \ref{tab:sample-num}, increasing $K$ from $16$ to $32$ does not improve the performance significantly. Therefore, in experiments related to \modelname{}, the number of proposed candidate retrieval results during testing is set to $k=16$.

Moreover, regarding the concern that introducing \modelname{} into the dialogue system may lead to unacceptable computational overhead, we report the time cost and maximum GPU memory usage during the training and inference with different retrieval models. From the results in Table \ref{tab:computation}, it can be seen that although the computational overhead of the dual-encoder based model is the smallest, the increase in computational overhead of the other retrievers is acceptable. Especially in scenarios such as vertical domain dialogues where the database size is relatively small, using more computational resources to improve the retrieval performance is preferable.

\vspace{-0.5em}

\section{Conclusion}

\vspace{-0.5em}

In this work, an energy-based retriever (\modelname{}) is proposed to collectively model the relevant knowledge pieces 
from a knowledge base given a context. 
\modelname{} can better model the inter-relationship between knowledge pieces, and can substantially improve the knowledge retrieval performance in knowledge-grounded dialog systems. Moreover, we conduct an in-depth exploration of various architectures of energy functions and training methods for \modelname{} and find out that using the residual form
can improve the quality of the retrieval results. 
Furthermore, in semi-supervised training of knowledge-grounded dialog systems, \modelname{} enables effective scoring of retrieved knowledge pieces, and leads to significant improvement in the end-to-end performance of dialog systems.
The above results show that \modelname{} has great potential for developing advanced knowledge-grounded dialog systems. 
We open-source the code and data to facilitate reproducibility and encourage further exploration in this direction. 
\vspace{-0.5em}
\section{Limitations}
\vspace{-0.5em}
In this work, training \modelname{}s with maximum likelihood estimate (MLE) methods is explored. However, in previous works, noise contrastive estimate (NCE) \cite{nce} methods have also been used to train energy-based language models. Therefore, training \modelname{}s with NCE methods can be studied in future works and compared with the MLE methods explored in this work. 

In addition, recent studies have explored using large language models (LLMs) for knowledge retrieval and reranking tasks. However, in this work, \modelname{} is implemented with relatively small models (BERT). Therefore, conducting experiments on \modelname{} with larger backbone models and studying the scaling effects of \modelname{} can be further explored. 

\bibliography{JSA}

\clearpage
\appendix
\section{The Details of the JSA Algorithm in Semi-Supervised Dialog Systems}
 The detailed procedure of semi-supervised training using the JSA algorithm is summarized in Algorithm \ref{alg:training}, which consists of two stages.
First, supervised pre-training is conducted on the retrieval model $p_\theta^{\text{ret}}$, the generation model $p_\theta^{\text{gen}}$, and the inference model $q_\phi$ on labeled data. 
After supervised pre-training, the retrieval parameters $\theta^{\text{ret}}$ are frozen in the second stage of training over unlabeled data.
Note that in unlabeled data, the knowledge pieces used in the dialogs are not annotated and the knowledge base (KB) is often not available.
This presents a significant challenge for the training of both the traditional retriever and the Entriever over unlabeled data.
Investigating the training of the retrieval parameters $\theta^{\text{ret}}$ over unlabeled data and unavailable KB is interesting future work.

In the second stage, supervised and unsupervised mini-batches are randomly drawn from labeled and unlabeled data. For labeled dialogs, the latent knowledge $\xi_t$ 
are given. For unlabeled dialogs, we apply the recursive turn-level MIS sampler based on Eq. (\ref{eq:tod-accept-reject}) to sample the latent states $\xi_t$ and treat them as if being given.
The gradient calculation and parameter updating are then the same for the labeled and unlabeled dialogs.
\needspace{100\baselineskip}
\label{sec:jsa}
\begin{algorithm}[h]
  \caption{JSA algorithm for training semi-supervised dialog systems}
  \label{alg:training}
  \begin{algorithmic}[1]
        \REQUIRE A mix of labeled and unlabeled dialogs.
	    \STATE Run supervised pre-training of $\theta=(\theta^{\text{ret}},\theta^{\text{gen}})$ and $\phi$ on labeled dialogs;
            \STATE Frozen the retriever parameters $\theta^{\text{ret}}$;
	    \REPEAT
	    \STATE Draw a dialog $(u_{1:T},r_{1:T})$;
		\IF  {$(u_{1:T},r_{1:T})$ is not labeled}
		\STATE Generate $\xi_{1:T}$ using the recursive turn-level MIS sampler;
		\ENDIF
		\STATE $J_{\theta^{\text{gen}}}=0, J_\phi=0$; 
        \FOR {$i = 1,\cdots,T$}
        \STATE $J_{\theta^{\text{gen}}} += \log p_{\theta}^{\text{gen}}(
        r_{t} \mid c_{t}, u_{t}, \xi_{t})$; \label{alg:line:p_gen}
        \STATE $J_\phi += \log q_\phi(\xi_{t} \mid c_{t}, u_{t}, r_{t})$;
        \ENDFOR
        \STATE Update $\theta^{\text{gen}}$ by ascending: $\nabla_{\theta^{\text{gen}}} J_{\theta^{\text{gen}}}$;
        \STATE Update $\phi$ by ascending: $\nabla_\phi J_\phi$;
        \UNTIL {convergence}
        \RETURN {$\theta$ and $\phi$}
  \end{algorithmic}
  \vspace{-0.4em}
\end{algorithm}


\end{document}